\relax
\documentclass[letterpaper]{article} 
\usepackage{custom}  
\usepackage{times}  
\usepackage{helvet}  
\usepackage{courier}  
\usepackage[hyphens]{url}  
\usepackage{graphicx} 
\usepackage{natbib}  
\usepackage{caption} 
\DeclareCaptionStyle{ruled}{labelfont=normalfont,labelsep=colon,strut=off} 
\usepackage{algorithm}
\usepackage{algorithmic}

\usepackage{newfloat}
\usepackage{listings}
\floatstyle{ruled}
\newfloat{listing}{tb}{lst}{}
\floatname{listing}{Listing}
\usepackage{amsmath} 
\usepackage{amsfonts}
\usepackage{bm} 
\usepackage{xstring} 
\usepackage{gensymb} 
\usepackage{booktabs} 
\usepackage{multirow} 
\usepackage{pifont} 
\usepackage{subcaption}
\graphicspath{{Figures/}}
\title{Learning the Physics of Particle Transport via Transformers}
\author{
    Oscar Pastor-Serrano, Zolt\'an Perk\'o
}
\affiliations{
    Delft University of Technology, \\
    Department of Radiation Science and Technology, \\
    Mekelweg 15 2629JB Delft, Netherlands \\
    o.pastorserrano@tudelft.nl
}
\begin{document}

\newcommand{\FixRef}[3][sec:]
{\IfBeginWith{#2}{#3}
	{\StrBehind{#2}{#3}[\RefResult]}
	{\def\RefResult{#2}}\IfBeginWith{#1}{#3}
	{\StrBehind{#1}{#3}[\RefResultb]}
	{\def\RefResultb{#1}}}

\newcommand{\secref}[1]
{\FixRef{#1}{sec:}Section~\ref{sec:\RefResult}}
\newcommand{\secreff}[1]
{\FixRef{#1}{sec:}in Section~\ref{sec:\RefResult}}
\newcommand{\Secreff}[1]
{\FixRef{#1}{sec:}In Section~\ref{sec:\RefResult}}
\newcommand{\secrefm}[2]
{\FixRef[#2]{#1}{sec:}Sections~\ref{sec:\RefResult}-\ref{sec:\RefResultb}}
\newcommand{\secreffm}[2]
{\FixRef[#2]{#1}{sec:}in Sections~\ref{sec:\RefResult}-\ref{sec:\RefResultb}}
\newcommand{\Secreffm}[2]
{\FixRef[#2]{#1}{sec:}In Sections~\ref{sec:\RefResult}-\ref{sec:\RefResultb}}
\newcommand{\figref}[1]
{\FixRef{#1}{fig:}Figure~\ref{fig:\RefResult}}
\newcommand{\figrefm}[2]
{\FixRef[#2]{#1}{fig:}Figures~\ref{fig:\RefResult}-\ref{fig:\RefResultb}}
\newcommand{\figreff}[1]
{\FixRef{#1}{fig:}in Figure~\ref{fig:\RefResult}}
\newcommand{\figreffm}[2
]{\FixRef[#2]{#1}{fig:}in Figures~\ref{fig:\RefResult}-\ref{fig:\RefResultb}}
\newcommand{\Figreff}[1]
{\FixRef{#1}{fig:}In Figure~\ref{fig:\RefResult}}
\newcommand{\Figreffm}[2]
{\FixRef[#2]{#1}{fig:}In Figures~\ref{fig:\RefResult}-\ref{fig:\RefResultb}}
\newcommand{\tabref}[1]
{\FixRef{#1}{tab:}Table~\ref{tab:\RefResult}}
\newcommand{\tabreff}[1]
{\FixRef{#1}{tab:}in Table~\ref{tab:\RefResult}}
\newcommand{\Tabreff}[1]
{\FixRef{#1}{tab:}In Table~\ref{tab:\RefResult}}
\newcommand{\tabrefm}[2]
{\FixRef[#2]{#1}{tab:}Tables~\ref{tab:\RefResult}-\ref{tab:\RefResultb}}
\newcommand{\tabreffm}[2]
{\FixRef[#2]{#1}{tab:}in Tables~\ref{tab:\RefResult}-\ref{tab:\RefResultb}}
\newcommand{\Tabreffm}[2]
{\FixRef[#2]{#1}{tab:}In Tables~\ref{tab:\RefResult}-\ref{tab:\RefResultb}}
\newcommand{\egyref}[1]
{\FixRef{#1}{eq:}Equation~\ref{eq:\RefResult}}
\newcommand{\eqreff}[1]
{\FixRef{#1}{eq:}in Equation~\ref{eq:\RefResult}}
\newcommand{\Eqreff}[1]
{\FixRef{#1}{eq:}In Equation~\ref{eq:\RefResult}}
\newcommand{\eqrefm}[2]
{Equations~\ref{eq:#1}-\ref{eq:#2}}
\newcommand{\eqreffm}[2]
{\FixRef[#2]{#1}{eq:}in Equations~\ref{eq:\RefResult}-\ref{eq:\RefResultb}}
\newcommand{\Eqreffm}[2]
{\FixRef[#2]{#1}{eq:}In Equations~\ref{eq:\RefResult}-\ref{eq:\RefResultb}}
\newcommand{\charef}[1]
{\FixRef{#1}{cha:}Chapter~\ref{cha:\RefResult}}
\newcommand{\chareff}[1]
{\FixRef{#1}{cha:}in Chapter~\ref{cha:\RefResult}}
\newcommand{\Chareff}[1]
{\FixRef{#1}{cha:}In Chapter~\ref{cha:\RefResult}}

\newcommand*{\dd}{\mathrm{d}}
\newcommand{\ui}[1]{\textit{\textbf{#1}}}
\newcommand{\mx}[1]{\underline{\underline{#1}}}
\newcommand{\diff}[2]{\dfrac{\dd #1}{\dd #2}}
\newcommand{\pdiff}[2]{\dfrac{\partial #1}{\partial #2}}
\newcommand{\dhl}{\hline\hline}
\newcommand{\rb}[1]{\left(#1\right)}
\newcommand{\sqb}[1]{\left[#1\right]}
\newcommand{\tb}[1]{\left<#1\right>}
\newcommand{\cb}[1]{\left\{#1\right\}}
\newcommand{\abs}[1]{\left|#1\right|}
\newcommand{\dspm}[1]{\begin{displaymath}#1\end{displaymath}}
\newcommand{\ds}{\displaystyle}
\newcommand{\pow}[2]{\cdot #1^{#2}}
\newcommand{\evat}[2]{\left.#1\right|_{#2}}
\newcommand{\ifrac}[2]{\ds #1 / #2}
\newcommand{\ab}{\ifrac{\alpha}{\beta}}
\newcommand{\BED}{\text{BED}}
\newcommand{\norm}[1]{\left\lVert#1\right\rVert}

\newcommand{\cmark}{\ding{51}}%
\newcommand{\xmark}{\ding{55}}%

\maketitle

\begin{abstract}
Particle physics simulations are the cornerstone of nuclear engineering applications. Among them radiotherapy (RT) is crucial for society, with 50\% of cancer patients receiving radiation treatments. For the most precise targeting of tumors, next generation RT treatments aim for real-time correction during radiation delivery, necessitating particle transport algorithms that yield precise dose distributions in sub-second times even in highly heterogeneous patient geometries. This is infeasible with currently available, purely physics based simulations. In this study, we present a data-driven dose calculation algorithm predicting the dose deposited by mono-energetic proton beams for arbitrary energies and patient geometries. Our approach frames particle transport as sequence modeling, where convolutional layers extract important spatial features into tokens and the transformer self-attention mechanism routes information between such tokens in the sequence and a beam energy token. We train our network and evaluate prediction accuracy using computationally expensive but accurate Monte Carlo (MC) simulations, considered the gold standard in particle physics. Our proposed model is 33 times faster than current clinical analytic pencil beam algorithms, improving upon their accuracy in the most heterogeneous and challenging geometries. With a relative error of $0.34\pm0.2$\% and very high gamma pass rate of $99.59\pm0.7$\% (1\%, 3 mm), it also greatly outperforms the only published similar data-driven proton dose algorithm, even at a finer grid resolution. Offering MC precision 400 times faster, our model could overcome a major obstacle that has so far prohibited real-time adaptive proton treatments and significantly increase cancer treatment efficacy. Its potential to model physics interactions of other particles could also boost heavy ion treatment planning procedures limited by the speed of traditional methods.
\end{abstract}

\section{Introduction}
\label{sec:Introduction}
Despite significant research efforts cancer remains a leading cause of death, responsible for more than 10 million deaths in 2020 worldwide \cite{GCO2021,Sung2021}. With more than 50\% of the patients receiving radiation treatments, radiotherapy (RT) is at the forefront of current standard of care, playing a crucially important role in improving societal health. Sophisticated computational methods and particle transport simulations have been key to this success \cite{Bernier2004}, enabling highly personalized treatments. Traditional physics based algorithms improved all steps in the RT workflow (imaging, segmentation, dose calculation, optimization), but so far they proved too slow and inaccurate for real-time adaptive treatments promising ultimate precision with fewest adverse side-effects. Deep learning is key to overcome these limitations and realize the full potential of real-time adaptation.

Our study focuses on learning particle transport physics --- fundamental to all steps of RT from Computed Tomography (CT) image reconstruction to simulating the actually delivered patient dose --- to provide the necessary sub-second speed and high accuracy required for real-time adaptation. We frame the transport problem as sequence modelling, with a particle beam going through varying geometries and materials, using convolutional layers to learn relevant spatial features and the transformer self-attention mechanism to combine information from the feature tokens and a beam energy token. We train our algorithm to specifically learn proton transport in lung cancer patients with highly heterogeneous geometries to predict dose based on CT images alone, but the model could in theory be easily adapted to other particles (photons, electrons, heavy ions) and quantities (e.g., particle flux or secondary particle emission prediction). 

\paragraph{Contributions} Our specific contributions are as follows:
\begin{itemize}
	\item We frame particle transport physics as a sequence modelling task and propose a novel algorithm using convolutional encoder and decoder layers together with transformer causal self-attention to predict dose distributions. 
	\item We train our algorithm using highly variable geometries from lung cancer patients and demonstrate that it outperforms both current clinical pencil beam algorithms (PBA), being 33 times faster and more precise in the most complex geometries, and 'gold standard' Monte Carlo (MC) methods, offering MC accuracy 400 times faster. Our model is also more accurate than the only published data driven proton dose calculation algorithm using (Long Short-Term Memory) LSTM cells.
	\item We highlight the direct societal impact of the presented algorithm by showcasing how it could improve current radiotherapy practice and enable real-time adaptive treatments. While we train our model to learn proton physics to predict dose distributions, we also detail extensions to make it a general particle transport simulator, accounting for e.g., beam shape or energy spectrum changes.
\end{itemize}

\section{Background}
\label{sec:Background}
Here we describe RT workflow and the critical role of particle transport and dose calculation methods, and frame our work in terms of unsolved challenges and related literature.

\subsubsection{Radiotherapy workflow}
RT treatments usually follow a 4-step procedure. First, high quality anatomical information is acquired --- typically as CT images \cite{Pereira2014} --- on which tumors to irradiate and organs at risk (OARs) to protect are delineated. Second, the irradiation modality is chosen, with most patients receiving photon treatments, but proton therapy spreading quickly due to protons' finite range and significantly better ability to focus dose on tumors \cite{Lundkvist2005}. Third, the beam angles and beamlet intensities to irradiate the patient with are optimized during treatment planning. This is the most complex and computationally expensive task, requiring solving large scale multi-criteria optimization problems and typically several iterations between planners and physicians before an acceptable, clinically 'best' plan is achieved \cite{Hussein2018,Meyer2018}. Last, for quality assurance purposes detailed dose calculations are performed to test plan robustness against anatomical changes or decide to adapt a plan for future irradiations.

\subsubsection{Particle transport \& dose calculation}
Accurate particle transport algorithms are crucial for all these steps. CT image reconstruction relies on simulating photon interactions with tissues and detectors; plan optimization requires the spatial dose distribution (typically in more than 1 million voxels) from each available proton or photon beamlet (in the thousands); while for plan evaluation the dose must be calculated for many different geometries. Ideally these calculations should be quick and precise, but current analytical pencil beam algorithms and stochastic MC dose calculation tools offer a trade-off. PBA yields results without the computational burden of MC engines, but its accuracy is severely compromised in highly heterogeneous or complex geometries, making slow and clinically rarely affordable MC approaches necessary. The problem is most acute for next generation real-time adaptive treatments promising  ultimate precision with fewest side effects by correcting treatments during irradiation, e.g., to account for anatomical changes due to breathing, coughs or intestinal movements. To finally become reality, such adaptive treatments require algorithms that deliver MC accuracy in sub-second speed.

\subsubsection{Related work}
Deep learning has achieved significant improvements in all steps of the RT workflow \cite{Meyer2018}, but only imaging, treatment planning and dose calculation are relevant to our work. U-net \cite{Ronneberger2015} and Generative Adversarial Networks \cite{Goodfellow2014} (and their variants) have been widely applied to improve image quality, e.g., to generate synthetic CT images from Magnetic Resonance Images (offering better soft tissue contrast than CT without additional patient dose) or low dose Cone-Beam CT (CBCT) images \cite{Edmund2017,Zhang2021}; to predict stopping power from CBCT \cite{Harms2020}; or correct scatter artifacts in CBCT reconstruction \cite{Lalonde2020}. These works represent image to image transformation, producing more useful images for the RT workflow than their easier/faster to obtain or lower patient dose input. 

In treatment planning, deep learning methods aim to predict an optimal 3D patient dose distribution achievable with a given radiotherapy technique based on historical data. The most successful works use ResNet based convolutional networks \cite{Chen2019,Fan2019}, 2/3D U-net \cite{Kearney2018,Nguyen2019, Kajikawa2019} or hierarchically densely connected U-net (HDU-net) \cite{Nguyen2019a,BarraganMontero2019} architectures, with segmented structure masks as input. Some also utilize the CT image \cite{Kearney2018} and manually encoded beam configuration information \cite{Nguyen2019a,BarraganMontero2019}. These works basically mimic 'optimal' plans for new patients that should be achievable based on past ones, only outputting final dose distributions, but not the required beam intensity (i.e., fluence) maps needed to deliver such plans, which must be obtained via additional, costly optimization. Thus, they are mostly used for Quality Assurance (QA) purposes to aid planning, not replace it. Only few papers attempt to jointly predict dose distribution and fluence maps \cite{Lee2019,Wang2020}, and all have been applied to photon treatments. 

Practically all applications of deep learning to dose calculations learn how to improve cheaper and faster physics based calculations. Most works try to predict low noise MC photon dose distributions from high noise MC doses \cite{Peng2019, Peng2019a,Bai2021,Neph2021}, or deterministic particle transport based photon distributions from simple analytical calculations \cite{Xing2020a,Dong2020}, using CNNs, U-net or HDU-net architectures with 2/3D patches. A few papers manually encode some physics information as additional input such as fluence maps \cite{Fan2020,Xing2020}, total energy released per unit mass maps \cite{Zhu2020} or beam information \cite{Kontaxis2020,Tsekas2021}. We are only aware of 2 papers \cite{Wu2021,Javaid2021} using deep learning to predict accurate low noise MC proton dose distributions, both using cheap physics models (noisy MC and PBA) as input. While all these works provide significant speed-up compared to pure physics based algorithms, some even reaching sub-second speeds, they all require physics models to produce their input, do not generalize easily (e.g., to different beam energies) and are trained with full plan data, unsuitable for real-time adaptation needing the individual dose distribution from each beamlet alone.

Most related to ours is the work from \cite{Neishabouri2021}, using LSTM networks to sequentially calculate proton pencil beam dose distributions from relative stopping power slices. Although requiring a separate model per beam energy, this LSTM-based dose engine offers excellent inference times and close to PBA accuracy when tested on external patient data. Our approach builds upon the methodology of \cite{Neishabouri2021}, but uses a different architecture, works on finer resolution and --- most crucially --- also learns the physics of energy dependence in particle transport via a single model.

\paragraph{Transformer} The backbone of the presented model is the Transformer, which was first introduced by \cite{Vaswani2017} for machine translation tasks. The Transformer and similar attention-based architectures have completely replaced recurrent neural network variants like LSTM in natural language processing applications since then \cite{Devlin2019, Brown2020}. A main reason behind their large-scale adoption and success is the ability to process long-term dependencies by directly accessing information at any point in the past without needing internal memory, which is essential to introduce beam energy dependence in our model.

Transformer-based architectures have also achieved state-of-the-art performance in computer vision tasks like image classification \cite{Ramachandran2019, Dosovitskiy2020}.  Inspired by \cite{Cordonnier2019}, \cite{Dosovitskiy2020} present the Vision Transformer (ViT), circumventing the quadratic cost of computing the attention weight matrix by dividing the input image into a patch sequence. Following the Transformer basis of ViT, our model also adopts a patch-based processing of the inputs.

The price that Transformers pay for their generality in both language and vision is the need for a self-supervised pre-training stage with large amounts of text or image data \cite{Devlin2019, Brown2020, Dosovitskiy2020}. For image classification, several approaches try to achieve state-of-the-art performance without costly pre-training \cite{Touvron2020, DAscoli2021}. As in the concurrent work of \cite{Hassani2021}, our model can be directly trained on a relatively small dataset by using a convolutional encoder first extracting important features from the patched input data.

\subsubsection{Self-attention} The proposed model leverages the self-attention (SA) mechanism \cite{Vaswani2017} allowing dynamic routing of information between the $L$ elements in a sequence $\bm{z}\in\mathbb{R}^{L\times D}$. SA is based on the interaction between a series of queries $\bm{Q}\in\mathbb{R}^{L\times D_h}$, keys $\bm{K}\in\mathbb{R}^{L\times D_h}$, and values $\bm{V}\in\mathbb{R}^{L\times D_h}$ obtained through a learned linear transformation of the input sequence 

\begin{equation}
	[\bm{Q}, \bm{K}, \bm{V}] = \bm{z}\bm{W}_{QKV},
\end{equation}

\noindent with learned weights $\bm{W}_{QKV}\in\mathbb{R}^{D\times 3D_h}$. Intuitively, every sequence element emits a query and a key vector with the information to gather from and to offer to the rest of the sequence, respectively. Each of the $L$ elements in the output sequence is a weighted sum of the values, where the weights --- referred to as attention matrix $\bm{A}\in\mathbb{R}^{L\times L}$ --- are obtained by matching queries against key vectors via inner products

\begin{equation}
	\bm{A} = \text{softmax}\Big(\frac{\bm{Q}\bm{K}^T}{\sqrt{D_h}}\Big),
\end{equation}

\begin{equation}
	\text{SA}(\bm{z})=\bm{A}\bm{V}.
\end{equation} 

Multi-head self-attention (MSA) runs $N_h$ parallel SA operations to extract different features and inter-dependencies, translating into setting $D_h=D/N_h$. The outputs of the different operations, called \textit{heads}, are concatenated and are linearly projected with learned weights $\bm{W}_h\in\mathbb{R}^{N_hD_h\times D}$ as

\begin{equation}
	\text{MSA}({\bm{z}}) = \underset{h\in\{N_h\}}{\text{concat}}[\text{SA}_h(\bm{z})]\bm{W}_h.
\end{equation}

By definition, MSA is invariant to the relative order of elements in the sequence. To account for positional information, fixed \cite{Vaswani2017} or learned \cite{Dosovitskiy2020} positional embedding can be added or concatenated to the input right before the first MSA operation. In addition, in MSA every element in the sequence can retrieve information at any past and future point. For some prediction tasks where the elements cannot or need not attend to future information, a binary mask is used to stop information flow from the future to the present. Such SA mechanism variant is referred to as causal SA and is particularly suited for modeling proton interactions and energy deposition physics that mostly occur sequentially in the forward beam direction.

\section{Methods}
\label{sec:Methods}
Our objective is to implicitly capture the physics of particle transport through a data-driven approach that enables accurate dose calculations in sub-second speed. The presented transformer-based parametric model exploiting the forward sequential nature of proton transport physics is well-suited for this. This section describes the model's building blocks, the dataset and the training and evaluation procedures.

\subsubsection{Proposed model} 
\begin{figure*}[t]
	\centering
	\includegraphics[width=0.9\textwidth]{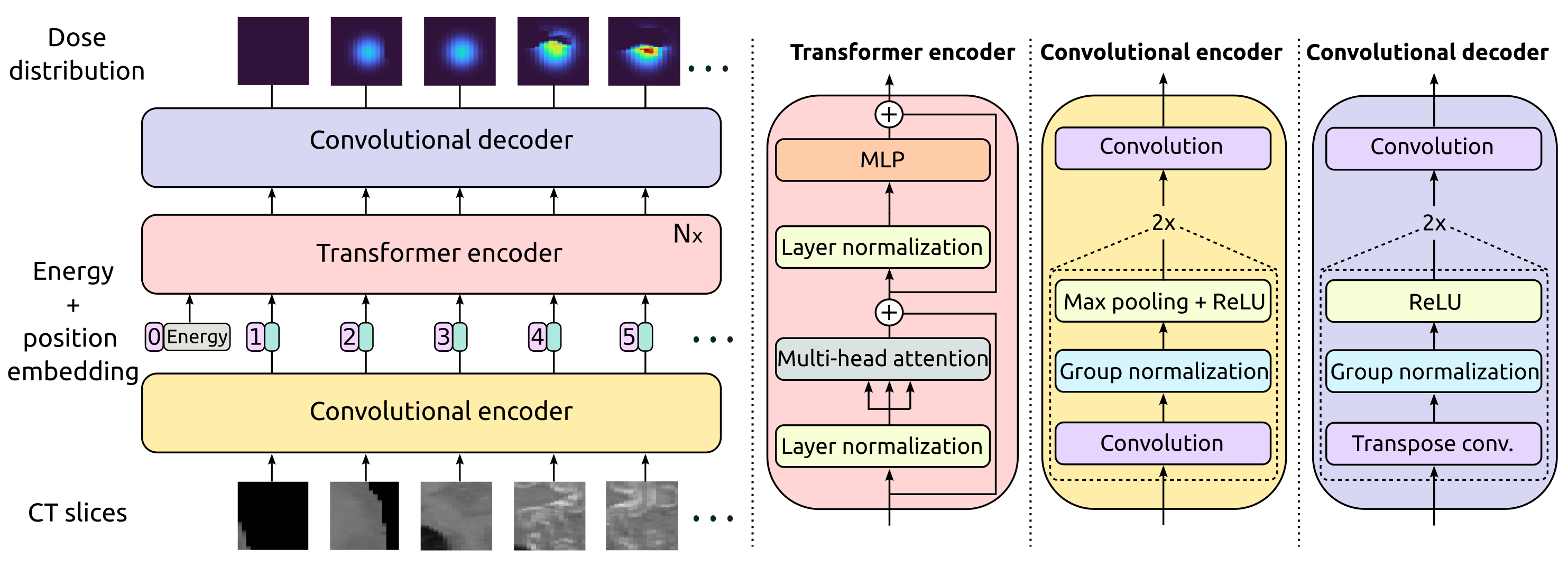} 
	\caption{Model architecture. We treat the input and output 3D volumes as a sequence of 2D slices. A convolutional encoder reduces the dimension of the input and extracts important geometrical features. The particle energy is added at the beginning of the resulting sequence. A transformer encoder with causal self-attention then routes information between the encoded input slices. Finally, a convolutional decoder transforms the low-dimensional sequence into an output sequence of 2D dose slices.}
	\label{fig:Architecture}
\end{figure*}

We introduce a parametric model that computes the output dose distribution $\bm{y}\in\mathbb{R}^{L\times H\times W}$ given the input geometry data $\bm{x}\in\mathbb{R}^{L\times H\times W}$ and particle energy $\epsilon\in E\subset\mathbb{R}^+$, where $L, H$ and $W$ are the depth, height and width of the geometric 3D grid, respectively. The model --- referred to as Dose Transformer Algorithm (DoTA) --- captures the relationship between the inputs and the output dose distribution through a nonlinear mapping $f_{\bm{\theta}}(\bm{x},\epsilon):\mathbb{R}^{L\times H\times W}\times E\rightarrow\mathbb{R}^{L\times H\times W}$, performed by a series of artificial neural networks. \figref{Architecture} shows the architecture of the model, which processes the 3D input geometry $\bm{x}$ as a sequence of $L$ 2D images in the beam's eye view $\{\bm{x}_i|\bm{x}_i\in\mathbb{R}^{1\times H\times W},\forall i=1,...,L \}$. 

\subsubsection{Convolutional encoder} First, a convolutional encoder extracts important features such as geometry contrasts and edges from the input CT slices. The convolutional encoder contains two blocks, each with a convolutional, a Group Normalization (GN) \cite{Wu2020} and a pooling layer, followed by a Rectified Linear Unit (ReLU) activation. After the second block, a convolution with $K$ filters results in a sequence of elements of reduced embedding dimension $D=H'\times W'\times K$, where $H'$ and $W'$ are the reduced height and width of the images. The layers in the convolutional encoder share weights and are applied independently to every element $\bm{x}_i$ in the sequence. We refer to the output of the convolutional encoder as tokens $\{\bm{z}_i|\bm{z}_i\in\mathbb{R}^{D},\forall i=1,...,L \}$.

\subsubsection{Transformer encoder} The interaction between tokens $\bm{z}_i$ is modeled in the transformer encoder through causal MSA, with each token routing information from all preceding tokens. To account for the relative positional information of sequence elements we add a learnable embedding to each token. We include an extra energy token $\bm{z}_e=\bm{W}_e\epsilon\in\mathbb{R}^D$ at the beginning of the sequence, where  $\bm{W}_e\in\mathbb{R}^{D\times 1}$ is a learned linear projection of the beam energy $\epsilon$. The transformer encoder alternates MSA and Multi-layer Perceptron (MLP) layers, with Layer Normalization (LN) \cite{Ba2016} and residual connections applied before and after every layer, respectively. A stack of $N$ transformer encoder blocks computes the operations

\begin{equation}
	\bm{z}_0 = [\bm{z}_e;\bm{z}] + \bm{r}_p,
\end{equation}
\begin{equation}
	\bm{s}_n = \bm{z}_{n-1} + \text{MSA}(\text{LN}(\bm{z}_{n-1})),   \qquad n=1...\:N
\end{equation}
\begin{equation}
	\bm{z}_n = \bm{s}_{n} + \text{MLP}(\text{LN}(\bm{s}_{n})), \quad\qquad n=1...\:N
\end{equation}

\noindent where $\bm{r}_p\in\mathbb{R}^{(L+1)\times D}$ is the learnable positional embedding and MLP is a two layer feed-forward network with Dropout \cite{Srivastava2014} and Gaussian Error Linear Unit (GELU) activations \cite{Hendrycks2016}.

\subsubsection{Convolutional decoder} To produce output dose volume $\bm{y}$ with the same dimension as the input, each token is transformed via a convolutional decoder with shared weights into the output slices $\{\bm{y}_i|\bm{y}_i\in\mathbb{R}^{1\times H\times W},\forall i=1,...,L \}$. Instead of normal convolutional layers, the decoder contains transposed convolutions that increase the dimension of their input. Similarly to the convolutional encoder, two final dimension-preserving convolutions transform the output of the second block into the 2D dose slices. 

\subsubsection{Dataset} The models are trained using a dataset with pairs of sliced CT images and dose distributions corresponding to mono-energetic proton beams with different energies. The 3D CT scans from 4 lung cancer patients are highly heterogeneous due to the air, bones and organs present in the thorax, and cover a volume of $512\times 512\times 100 $ $\text{mm}^3$ with a resolution of $1\times 1\times 3$ mm. Since each proton beam has approximately 20 mm diameter and travels up to 250 mm through a small volume only, we crop and extract blocks $\bm{x}\in\mathbb{R}^{256\times 48\times 16}$ maintaining the original CT resolution. Many different blocks can be obtained from the same patient by rotating the CT scan along the Z direction in steps of $5\degree$ and applying shifts in YZ plane with $5$ mm steps.

The output ground-truth dose distributions are calculated using the open source Monte Carlo particle transport software MCsquare \cite{Souris2016}, taking CT slices and calculating output blocks $\bm{y}\in\mathbb{R}^{256\times 48\times 16}$ with the same size and resolution as the input via random sampling of proton trajectories. Dose distributions are estimated using 3 million primary particles ensuring low MC noise levels of 0.6\%. For each input CT block we generate 4 dose distributions corresponding to 4 randomly sampled energies between 80 and 130~MeV, rounded to 1 decimal. We mask MC noise by zeroing out dose values below the noise level.

The entire training dataset consists of 63,048 pairs of input-output blocks, 10\% of which are used as a validation set. We apply data augmentation during training and randomly rotate the volumes $180\degree$ in beam's eye view (YZ plane), doubling the number of samples. A test set of 3,618 input-output pairs from an external patient is used to evaluate generalization to unseen geometries and energies.

\subsubsection{Training details}
The best performing model consists of one transformer encoder block with 16 heads and convolutional layers with a 3$\times$3 kernel. Using size preserving zero-padding results in halving (or doubling, for the decoder) the H and W dimensions after each convolutional block. The token embedding dimension $D=H/4\times W/4\times K$ is constant throughout the transformer encoder layers, with height $H=48$, width $W=16$, $K=10$ kernels and $D=480$ in our particular case. The models are trained with Tensorflow \cite{TF2015} using the LAMB optimizer \cite{You2019} and mini-batches of 8 samples, limited by the maximum internal memory of the Nvidia Tesla T4{\textregistered} Graphics Processing Unit (GPU) used during our experiments. We use the mean squared error (MSE) as loss function and a scheduled learning rate starting at $10^{-3}$ that is halved every 4 epochs. In Appendix A we perform a model hyperparameter search varying the number of transformer layers $N$, convolutional filters $K$ and attention heads $N_h$.

\subsubsection{Gamma analysis}
We compare the predicted and ground-truth 3D dose distributions from the test set using gamma analysis \cite{Low1998}.  Intuitively, for a set reference points and their corresponding reference dose values, this method searches for similar dose values within small spheres around each point. The similarity is quantified using a maximum dose difference threshold (usually expressed as a percentage of the reference dose): e.g., dose values are accepted similar if within 1\% of the reference dose. The radius of the sphere is referred to as distance-to-agreement criterion. Mathematically, gamma values are calculated for individual points in the predicted dose grid as 

\begin{equation}
	\gamma(\bm{p}) = \underset{\bm{\hat{p}}}{\min}\{\Gamma(\bm{p},\bm{\hat{p}})\},
\end{equation}

\begin{equation}
	\Gamma_{\delta, \Delta}(\bm{p},\bm{\hat{p}})=\sqrt{\frac{\abs{\bm{p}-\bm{\hat{p}}}^2}{\delta^2}+\frac{\abs{D(\bm{p})-D(\bm{\hat{p}})}^2}{\Delta^2}},
\end{equation}

\noindent where $\bm{p}$ and $\bm{\hat{p}}$ are the coordinates of the points in the predicted and ground truth dose grids, respectively. $D(\bm{p})$ is the dose at any point $\bm{p}$, $\delta$ is the distance-to-agreement and $\Delta$ the dose difference criterion.

We use the publicly available gamma evaluation functions from PyMedPhys\footnote{see \url{https://docs.pymedphys.com}}, with $\delta=3$ mm and $\Delta=1\%$. The 3 mm distance-to-agreement criterion ensures a neighborhood search of at least one voxel, while the dose difference criterion of 1\% disregards uncertainty due to MC noise. Gamma values are calculated for each voxel and a voxel centered at $\bm{p}$ is considered to pass the gamma evaluation if $\gamma(\bm{p})<1$. For the entire grid, the gamma pass rate can be calculated as the fraction of passed voxels over total number of voxels.

\subsubsection{Error analysis} The sample average relative error is used as an additional method to explicitly compare dose differences between two grids. Given the predicted output $\bm{y}$ and the ground truth dose distribution $\bm{\hat{y}}$, the average relative error $\rho$ can be calculated as

\begin{equation}
	\rho = \frac{\norm{\bm{y}-\bm{\hat{y}}}_{L_1}}{\max_j{\bm{\hat{y}}_j}}\times 100,
\end{equation}

\noindent where $\max_j{\bm{\hat{y}}_j}$ is the maximum dose value among all voxels in the ground-truth dose grid.

\section{Experiments}
\label{sec:Experiments}

We compare the performance of the presented DoTA model to both state-of-the-art and clinically used methods. The experiments first focus on evaluating the accuracy all models: the gamma evaluation serves as a tool to assess dosimetric differences, while the relative error allows direct comparison of the predicted output and ground truth grids. Last, we report calculation times and evaluate DoTAs' potential to displace other algorithms as a fast dose calculation tool.

\subsubsection{Baselines} Our approach is compared to PBAs, the group of analytical dose calculation methods mostly used in the clinic. In particular, we calculate dose distributions for the entire test set using the PBA included in the open-source treatment planning software matRad\footnote{Available at \url{https://github.com/e0404/matRad}.} \cite{Wieser2017}. The DoTA model is also compared to the only published data-driven approach based on LSTM cells \cite{Neishabouri2021}. Since the LSTM models in \cite{Neishabouri2021} are trained for a single energy, we additionally train a model using 104.25 megaelectronvolt  (MeV) proton beams. 

\subsubsection{Gamma evaluation} The gamma pass rate is calculated for every test sample using the MC dose distributions as reference with two settings. In the first \textit{unmasked} setting identical to \cite{Neishabouri2021}, voxels with exactly 0 gamma value are excluded from the pass rate calculation. These are typically voxels not receiving any dose in both the predicted and ground-truth grids, having no clinical relevance. However, the outputs of the model's last linear layer are hardly ever exactly 0, taking very small values instead. Thus, in the second, stricter \textit{masked} setting, we mask voxels in the predicted dose grid that are below 0.01\% of the maximum dose.

\begin{figure}[bth!]
	\centering
	\includegraphics[width=0.45\textwidth]{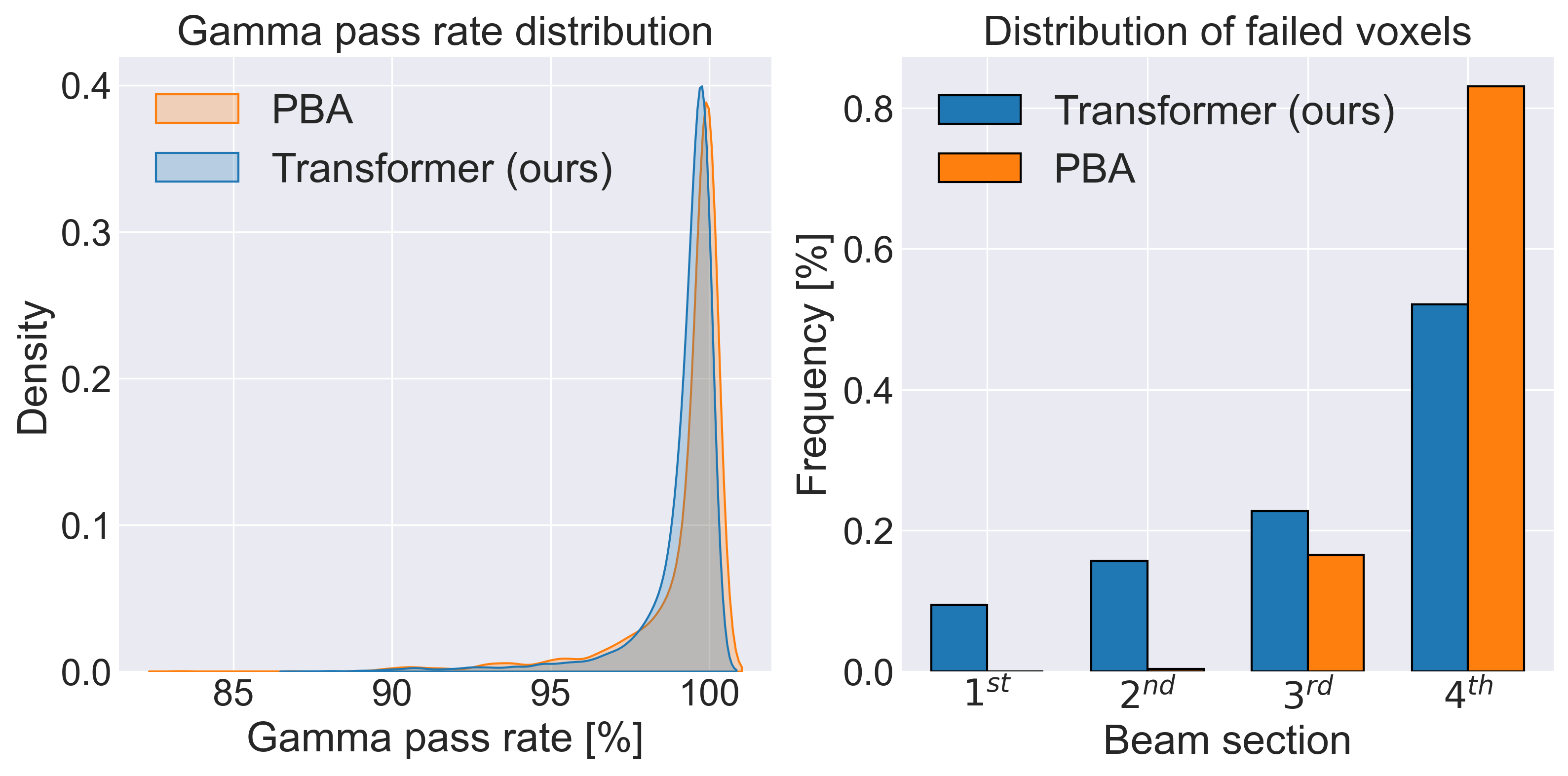}
	\caption{(Left) Distribution of the gamma pass rates across the test set for the PBA and DoTA. (Right) Distribution of the failed voxels along the beam, where each bin shows the ratio between the number of voxels in the test set that fail the gamma evaluation within a section of the beam and the total number of failed voxels.}
	\label{fig:gpr}
\end{figure}

\tabref{gpr} summarizes the results of the gamma evaluation for both settings. We report mean, standard deviation, maximum and minimum pass rates across the entire test set. Even with energy dependence and a finer grid resolution, DoTA outperforms the LSTM model in all aspects: the average pass rate is higher, the spread lower, and the minimum is almost 2\% higher. The performance of PBA and DoTA is very similar: their average values are very close in both the masked and unmasked setting, and their gamma pass rate distributions (left plot of \figref{gpr}) almost overlap. The minimum pass rate is significantly higher for DoTA, indicating that PBA struggles with the most heterogeneous and complicated samples. To verify this, we divide the dose beamlet into 4 equal sections along depth and score the number of failed voxels per section across the entire test set. The right plot \figreff{gpr} shows the proportion of voxels failing the gamma evaluation per section, out of the total number of failed voxels. The higher proportion in the $4^{\text{th}}$, last section of the beam represents inaccuracies in the high dose, clinically most relevant regions where the effects of heterogeneities are most evident.

\begin{table*}[]
	\centering
	\caption{Gamma analysis results ($\delta=3 \text{mm}$, $\Delta=1\%$). For PBA and DoTA, gamma pass rates are calculated across the same test set. Pass rates of the LSTM models are directly obtained from \cite{Neishabouri2021}. Mask indicates whether the predicted low dose values below 0.01\% of the maximum dose are masked before gamma evaluation. Mean, standard deviation (Std), minimum (Min) and maximum (Max) values across the test set are shown for each model, mask and energy combination.}
	\begin{tabular}{@{}lcccccc@{}}
		\toprule
		\textbf{Model} & \textbf{Mask} & \textbf{Energy (MeV)} & \textbf{Mean (\%)} & \textbf{Std (\%)} & \textbf{Min (\%)} & \textbf{Max (\%)} \\ \midrule
		\multirow{3}{*}{LSTM \cite{Neishabouri2021}} & \xmark & 67.85 & 98.56 & 1.30 & 95.35 & 99.79 \\
		& \xmark & 104.25 & 97.74 & 1.48 & 92.57 & 99.74 \\
		& \xmark & 134.68 & 94.51 & 2.99 & 85.37 & 99.02 \\ \midrule
		\multirow{3}{*}{DoTA (ours)} & \xmark & 104.25 & 99.55 & 0.71 & 93.45 & 100 \\
		& \xmark & [80, 130] & 99.59 & 0.70 & 92.79 & 100 \\
		& \cmark & [80, 130] & 99.12 & 1.45 & 87.32 & 100 \\ \midrule
		\multirow{2}{*}{PBA \cite{Wieser2017}} & \xmark & [80, 130] & 99.45 & 1.16 & 89.61 & 100 \\
		& \cmark & [80, 130] & 99.16 & 1.73 & 83.35 & 100 \\ \bottomrule
	\end{tabular}
\label{tab:gpr}
\end{table*}

\subsubsection{Error evaluation} 
\begin{figure}[]
	\centering
	\includegraphics[width=0.45\textwidth]{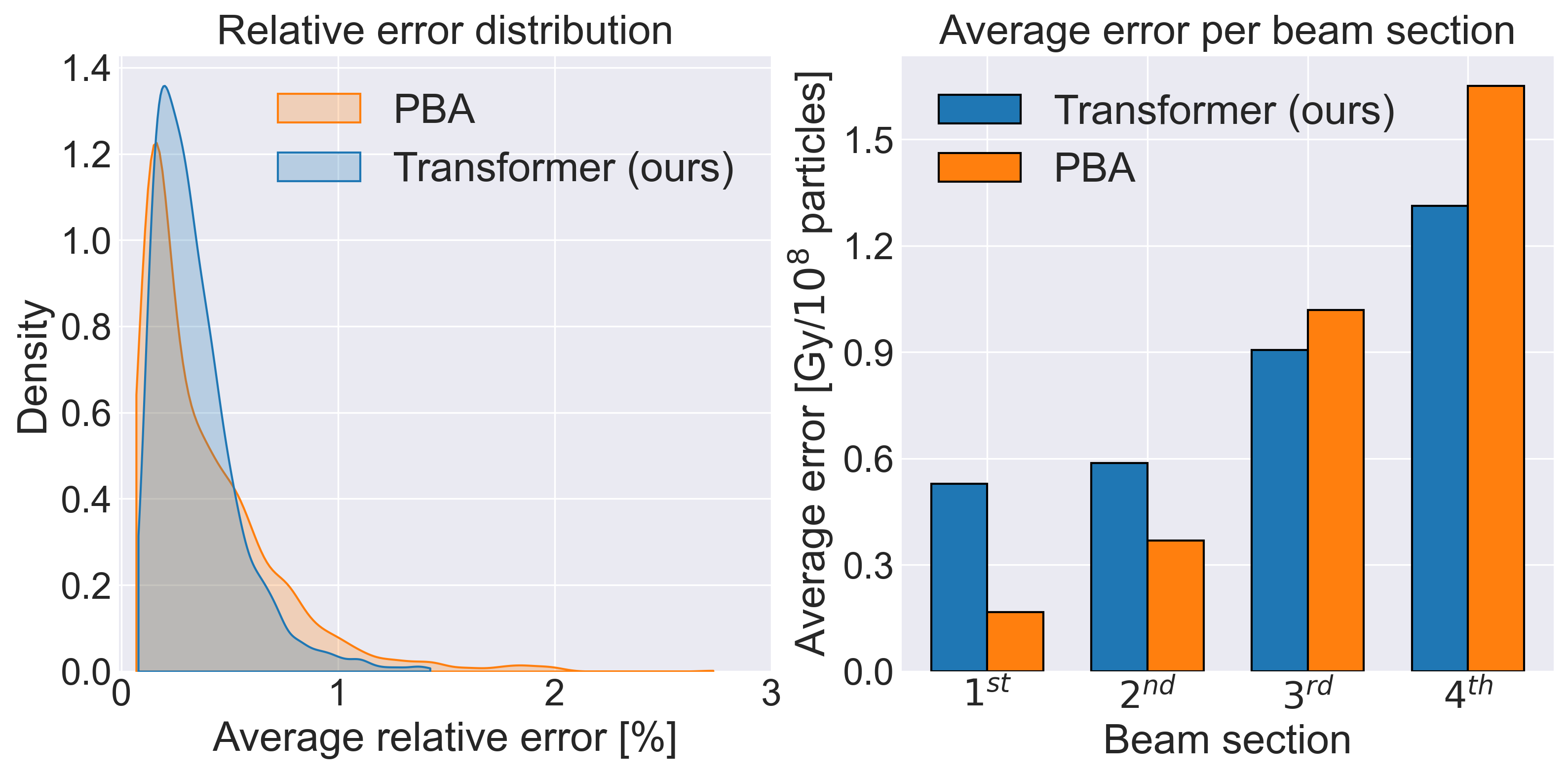}
	\caption{(Left) Distribution of the average relative errors across the test set for the PBA and DoTA. (Right) Comparison of the average relative error per beam section, where each bin shows the mean relative error  across the test set for equally sized sections of the beam.}
	\label{fig:err}
\end{figure}

To explicitly compare the performance of PBA and DoTA, we calculate the sample average relative error $\rho$ of the test set. \tabref{err} shows the mean, standard deviation, maximum and minimum errors observed across all test samples, and the left plot in \figref{err} displays the distribution of $\rho$ values for both models. Though PBA achieves low errors in the most homogeneous samples, our approach is clearly superior with a lower mean $\rho$ and a twice lower maximum error. The depth profile in the right plot of \figref{err} shows the same trend as the gamma evaluation: DoTA outperforms PBA in the last high dose regions of the beam.

\begin{table}[h!]
	\centering
	\caption{Average relative error between predicted and reference MC dose distributions. Reported values include mean, standard deviation (Std), minimum (Min) and maximum (Max) values across the test set, for both the PBA and DoTA.}
	\begin{tabular}{@{}lllll@{}}
		\toprule
		\textbf{Model} & \textbf{Mean (\%)} & \textbf{Std (\%)} & \textbf{Min (\%)} & \textbf{Max (\%)} \\ \midrule
		\begin{tabular}[c]{@{}l@{}}DoTA\\ (ours)\end{tabular} & 0.3430 & 0.1999 & 0.0780 & 1.4250 \\
		PBA & 0.3863 & 0.3154 & 0.0683 & 2.7317 \\ \bottomrule
	\end{tabular}
	\label{tab:err}
\end{table}

\subsubsection{Time evaluation} Besides high prediction accuracy, fast inference times are critical for clinical dose calculation algorithms. \tabref{time} reports run times of the LSTM, DoTA, PBA and MC dose algorithms. Though dependent on hardware, the data-driven approaches are clearly faster than clinically used PBA and MC baselines. While LSTM seems faster than DoTA, this is partially due to the LSTM model having 2.67 times smaller input/output and being run on better hardware. Our proposed approach offers a 33 and 400 times speed-up compared to the PBA and MC methods, respectively.

\begin{table}[h!]
	\caption{Mean inference time and standard deviation (Std) across the test set for each model. Reported run times only account for the dose calculation and disregard pre-processing steps. The values for the LSTM model are taken directly from \cite{Neishabouri2021}. The DoTA runtimes include the per-sample runtime obtained using the maximum GPU capacity corresponding to a batches of 8 sample.}
	\begin{center}
		\begin{tabular}{@{}lll@{}}
			\toprule
			\textbf{Model} & \textbf{Mean (ms)} & \textbf{Std (ms)} \\ \midrule
			LSTM$^a$ & 6.0 & - \\
			DoTA$^b$ (1 per batch) & 70.5 & 10.2 \\
			DoTA$^b$ (8 per batch) & 31.2 & 1.0 \\
			PBA$^c$ & 1,030.7 & 108.9 \\
			MC$^c$ & 13,295.2 & 3,180.0 \\ \bottomrule
		\end{tabular}\\				
	\end{center}
	\label{tab:time}
	\footnotesize{$^a$ Nvidia{\small\textregistered} Quadro RTX 6000 64 Gb RAM}\\
	\footnotesize{$^b$ Debian20 4 vCPUs 15 Gb RAM - Nvidia{\small\textregistered} Tesla T4 16 Gb RAM}\\
	\footnotesize{$^c$ Ubuntu 20.04 intel{\small\textregistered} Core\textsuperscript{\tiny TM}} i7-8550U 1.8 GHz 16Gb RAM
\end{table}

\begin{figure*}[h!]
	\centering
	\begin{subfigure}[b]{0.49\textwidth}
		\centering
		\includegraphics[width=\textwidth]{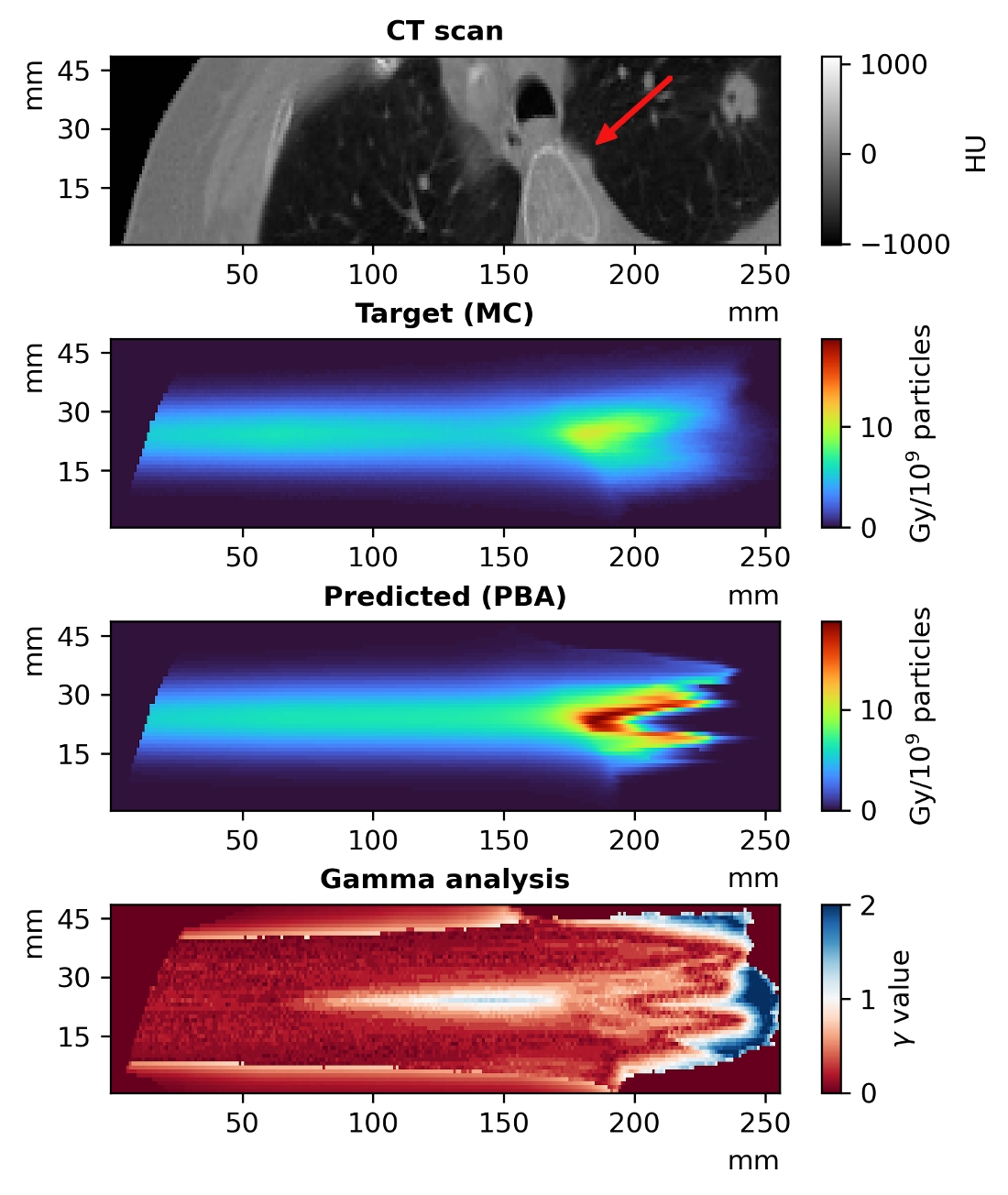}
		\caption{PBA}
		\label{fig:worst_pba}
	\end{subfigure}
	\begin{subfigure}[b]{0.49\textwidth}
		\centering
		\includegraphics[width=\textwidth]{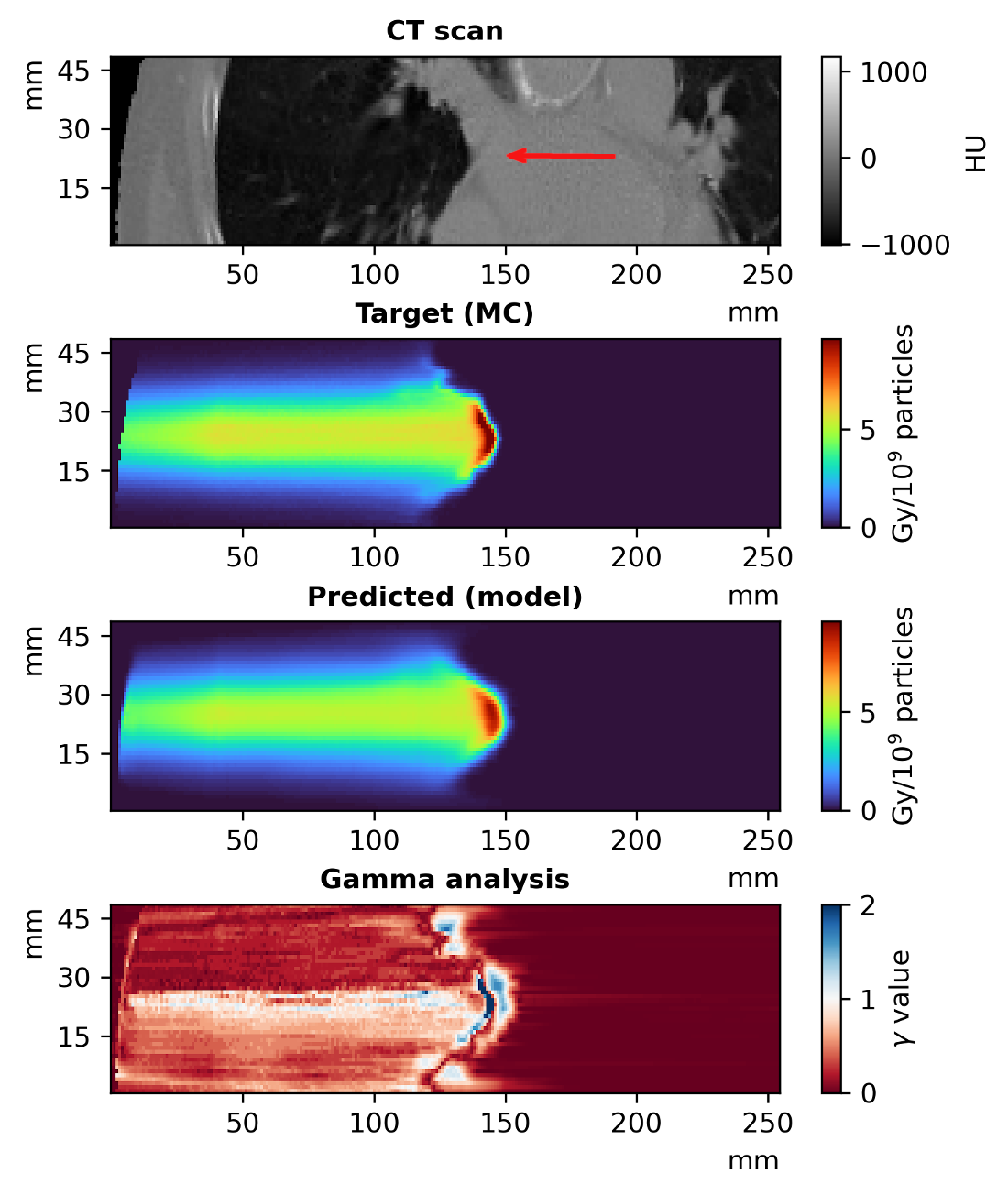}
		\caption{DoTA}
		\label{fig:worst_dota}
	\end{subfigure}
	\caption{Worst performing sample in the gamma evaluation across the test set, for (a) PBA and (b) DoTA. Each plot displays the central slice of the 3D input CT grid, the MC ground truth dose distribution, the model prediction and the gamma values.}
	\label{fig:worst}
\end{figure*}

\subsubsection{Additional geometries} \figref{worst} displays the worst performing test sample in terms of gamma pass rate for the PBA and DoTA models. Both samples consist of a beam traversing the lungs, where most of the energy is deposited in a highly heterogeneous region, with the PBA sample completely misplacing the high energy peak, while DoTA still yielding reasonable prediction. In Appendix B we evaluate the model in additional geometries unseen during training.

\section{Conclusions}
\label{sec:Conclusions}
After their recent success in natural language processing and computer vision tasks, transformer-based architectures prove to excel in problems that involve sequential image processing too. Framing particle transport as sequence modeling of 2D geometry slices, we use their power to build a fast and accurate dose calculation algorithm that implicitly learns proton transport physics and has the potential for profound social impact by enabling next generation, real-time adaptive radiotherapy cancer treatments.

Our evaluation shows that the presented DoTA model has the right attributes to replace proton dose calculation algorithms currently used in clinical practice. Compared to PBAs, DoTA achieves 33x faster inference times while being better suited for heterogeneous patient geometries. The high gamma pass rate in unseen geometries from an external patient also demonstrate that the model predicts close to high accuracy MC dose distributions in sub-second times. 

Such speed and accuracy increase could directly improve current RT practice by allowing comprehensive plan robustness analysis (now performed by checking only few potential geometries), quick dosimetric quality assurance of daily treatments (mostly done by analysing anatomy changes via comparing pre-treatment CT/CBCT images to the planning CT instead of corresponding dose distributions variations) and precise evaluation of needing plan adaptation. Crucially, the sub-second speed for individual pencil beam dose calculation and incorporation of energy dependence make our model well suited for real-time treatment adaptation. 

To our knowledge DoTA is the first deep learning method to implicitly learn particle transport physics, predicting dose using only CT and beam energy as input, as opposed to previous works only learning corrections for 'cheap' physics based predictions or predicting dose under fixed conditions. The flexibility to incorporate additional beam characteristics, e.g., changes in beam shape (provided as 0th image slice) or energy spectrum (as the $\epsilon$ 0th token), or to predict additional quantities (e.g., particle flux) holds the potential to be a fast, general particle transport algorithm. Since photons (used in photon therapy and CT/CBCT imaging) and heavy ions (in carbon and helium therapy) share similar, mostly forward scatter physics, training our algorithm for different particles could open door to several further applications: e.g., predicting physical or radiobiological dose (requiring DNA scale simulations) in heavy ion treatments; or real-time CBCT image reconstruction to provide input for real-time adaptation. Attending to future information too could even allow learning large angle scatter physics crucial for electron therapy. Thus, the presented algorithm could significantly contribute to improving cancer treatments, having profound societal impact even on the short term.

\section{Acknowledgments}
The authors would like to thank Kevin Wielinga for his contributions to this project. This work is supported by KWF Kanker Bestrijding [grant number 11711], and is part of the KWF research project PAREL. Zolt\'an Perk\'o would like to thank the support of the NWO VENI grant ALLEGRO (016.Veni.198.055) during the time of this study. 

\section*{Code availability}
The code, weights and results are publicly available at \url{https://github.com/} (to be released after the review process).

\section{CRediT authorship contribution statement.}
\textbf{Oscar Pastor-Serrano}: Conceptualization, Methodology, Software, Validation, Formal Analysis, Investigation, Data Curation, Writing – original draft, Visualization.

\noindent \textbf{Zolt\'an Perk\'o}: Conceptualization, Methodology, Formal Analysis, Resources, Writing – original draft, Writing – Review \& editing, Supervision, Project Administration, Funding Acquisition.

\small
\bibliography{bibliography}

\begin{thebibliography}{54}
\providecommand{\natexlab}[1]{#1}

\bibitem[{Abadi et~al.(2015)Abadi, Agarwal, Barham, Brevdo, Chen, Citro,
  Corrado, Davis, Dean, Devin, Ghemawat, Goodfellow, Harp, Irving, Isard, Jia,
  Jozefowicz, Kaiser, Kudlur, Levenberg, Man\'{e}, Monga, Moore, Murray, Olah,
  Schuster, Shlens, Steiner, Sutskever, Talwar, Tucker, Vanhoucke, Vasudevan,
  Vi\'{e}gas, Vinyals, Warden, Wattenberg, Wicke, Yu, and Zheng}]{TF2015}
Abadi, M.; Agarwal, A.; Barham, P.; Brevdo, E.; Chen, Z.; Citro, C.; Corrado,
  G.~S.; Davis, A.; Dean, J.; Devin, M.; Ghemawat, S.; Goodfellow, I.; Harp,
  A.; Irving, G.; Isard, M.; Jia, Y.; Jozefowicz, R.; Kaiser, L.; Kudlur, M.;
  Levenberg, J.; Man\'{e}, D.; Monga, R.; Moore, S.; Murray, D.; Olah, C.;
  Schuster, M.; Shlens, J.; Steiner, B.; Sutskever, I.; Talwar, K.; Tucker, P.;
  Vanhoucke, V.; Vasudevan, V.; Vi\'{e}gas, F.; Vinyals, O.; Warden, P.;
  Wattenberg, M.; Wicke, M.; Yu, Y.; and Zheng, X. 2015.
\newblock {TensorFlow}: Large-Scale Machine Learning on Heterogeneous Systems.
\newblock Software available from tensorflow.org.

\bibitem[{Ba, Kiros, and Hinton(2016)}]{Ba2016}
Ba, J.~L.; Kiros, J.~R.; and Hinton, G.~E. 2016.
\newblock {Layer Normalization}.

\bibitem[{Bai et~al.(2021)Bai, Wang, Nguyen, and Jiang}]{Bai2021}
Bai, T.; Wang, B.; Nguyen, D.; and Jiang, S. 2021.
\newblock {Deep dose plugin: towards real-time Monte Carlo dose calculation
  through a deep learning-based denoising algorithm}.
\newblock \emph{Machine Learning: Science and Technology}, 2(2): 025033.

\bibitem[{Barrag{\'{a}}n‐Montero et~al.(2019)Barrag{\'{a}}n‐Montero,
  Nguyen, Lu, Lin, Norouzi‐Kandalan, Geets, Sterpin, and
  Jiang}]{BarraganMontero2019}
Barrag{\'{a}}n‐Montero, A.~M.; Nguyen, D.; Lu, W.; Lin, M.-H.;
  Norouzi‐Kandalan, R.; Geets, X.; Sterpin, E.; and Jiang, S. 2019.
\newblock {Three‐dimensional dose prediction for lung IMRT patients with deep
  neural networks: robust learning from heterogeneous beam configurations}.
\newblock \emph{Medical Physics}, 46(8): 3679--3691.

\bibitem[{Bernier, Hall, and Giaccia(2004)}]{Bernier2004}
Bernier, J.; Hall, E.~J.; and Giaccia, A. 2004.
\newblock {Radiation oncology: a century of achievements}.
\newblock \emph{Nature Reviews Cancer}, 4(9): 737--747.

\bibitem[{Brown et~al.(2020)Brown, Mann, Ryder, Subbiah, Kaplan, Dhariwal,
  Neelakantan, Shyam, Sastry, Askell, Agarwal, Herbert-Voss, Krueger, Henighan,
  Child, Ramesh, Ziegler, Wu, Winter, Hesse, Chen, Sigler, Litwin, Gray, Chess,
  Clark, Berner, McCandlish, Radford, Sutskever, and Amodei}]{Brown2020}
Brown, T.~B.; Mann, B.; Ryder, N.; Subbiah, M.; Kaplan, J.; Dhariwal, P.;
  Neelakantan, A.; Shyam, P.; Sastry, G.; Askell, A.; Agarwal, S.;
  Herbert-Voss, A.; Krueger, G.; Henighan, T.; Child, R.; Ramesh, A.; Ziegler,
  D.~M.; Wu, J.; Winter, C.; Hesse, C.; Chen, M.; Sigler, E.; Litwin, M.; Gray,
  S.; Chess, B.; Clark, J.; Berner, C.; McCandlish, S.; Radford, A.; Sutskever,
  I.; and Amodei, D. 2020.
\newblock {Language models are few-shot learners}.
\newblock \emph{Advances in Neural Information Processing Systems}, 2020-Decem.

\bibitem[{Chen et~al.(2019)Chen, Men, Li, Yi, and Dai}]{Chen2019}
Chen, X.; Men, K.; Li, Y.; Yi, J.; and Dai, J. 2019.
\newblock {A feasibility study on an automated method to generate
  patient-specific dose distributions for radiotherapy using deep learning}.
\newblock \emph{Medical Physics}, 46(1): 56--64.

\bibitem[{Cordonnier, Loukas, and Jaggi(2019)}]{Cordonnier2019}
Cordonnier, J.-B.; Loukas, A.; and Jaggi, M. 2019.
\newblock {On the Relationship between Self-Attention and Convolutional
  Layers}.

\bibitem[{D'Ascoli et~al.(2021)D'Ascoli, Touvron, Leavitt, Morcos, Biroli, and
  Sagun}]{DAscoli2021}
D'Ascoli, S.; Touvron, H.; Leavitt, M.; Morcos, A.; Biroli, G.; and Sagun, L.
  2021.
\newblock {ConViT: Improving Vision Transformers with Soft Convolutional
  Inductive Biases}.

\bibitem[{Devlin et~al.(2019)Devlin, Chang, Lee, and Toutanova}]{Devlin2019}
Devlin, J.; Chang, M.~W.; Lee, K.; and Toutanova, K. 2019.
\newblock {BERT: Pre-training of deep bidirectional transformers for language
  understanding}.
\newblock \emph{NAACL HLT 2019 - 2019 Conference of the North American Chapter
  of the Association for Computational Linguistics: Human Language Technologies
  - Proceedings of the Conference}, 1(Mlm): 4171--4186.

\bibitem[{Dong and Xing(2020)}]{Dong2020}
Dong, P.; and Xing, L. 2020.
\newblock {Deep DoseNet: a deep neural network for accurate dosimetric
  transformation between different spatial resolutions and/or different dose
  calculation algorithms for precision radiation therapy}.
\newblock \emph{Physics in Medicine \& Biology}, 65(3): 035010.

\bibitem[{Dosovitskiy et~al.(2020)Dosovitskiy, Beyer, Kolesnikov, Weissenborn,
  Zhai, Unterthiner, Dehghani, Minderer, Heigold, Gelly, Uszkoreit, and
  Houlsby}]{Dosovitskiy2020}
Dosovitskiy, A.; Beyer, L.; Kolesnikov, A.; Weissenborn, D.; Zhai, X.;
  Unterthiner, T.; Dehghani, M.; Minderer, M.; Heigold, G.; Gelly, S.;
  Uszkoreit, J.; and Houlsby, N. 2020.
\newblock {An Image is Worth 16x16 Words: Transformers for Image Recognition at
  Scale}.

\bibitem[{Edmund and Nyholm(2017)}]{Edmund2017}
Edmund, J.~M.; and Nyholm, T. 2017.
\newblock {A review of substitute CT generation for MRI-only radiation
  therapy.}
\newblock \emph{Radiation oncology (London, England)}, 12(1): 28.

\bibitem[{Fan et~al.(2019)Fan, Wang, Chen, Hu, Zhang, and Hu}]{Fan2019}
Fan, J.; Wang, J.; Chen, Z.; Hu, C.; Zhang, Z.; and Hu, W. 2019.
\newblock {Automatic treatment planning based on three-dimensional dose
  distribution predicted from deep learning technique}.
\newblock \emph{Medical Physics}, 46(1): 370--381.

\bibitem[{Fan et~al.(2020)Fan, Xing, Dong, Wang, Hu, and Yang}]{Fan2020}
Fan, J.; Xing, L.; Dong, P.; Wang, J.; Hu, W.; and Yang, Y. 2020.
\newblock {Data-driven dose calculation algorithm based on deep U-Net}.
\newblock \emph{Physics in Medicine \& Biology}, 65(24): 245035.

\bibitem[{Ferlay et~al.(2020)Ferlay, Ervik, Lam, Colombet, Mery, Pi{\~{n}}eros,
  Znaor, Soerjomataram, and Bray}]{GCO2021}
Ferlay, J.; Ervik, M.; Lam, F.; Colombet, M.; Mery, L.; Pi{\~{n}}eros, M.;
  Znaor, A.; Soerjomataram, I.; and Bray, F. 2020.
\newblock Global Cancer Observatory: Cancer Today.

\bibitem[{Goodfellow et~al.(2014)Goodfellow, Pouget-Abadie, Mirza, Xu,
  Warde-Farley, Ozair, Courville, and Bengio}]{Goodfellow2014}
Goodfellow, I.; Pouget-Abadie, J.; Mirza, M.; Xu, B.; Warde-Farley, D.; Ozair,
  S.; Courville, A.; and Bengio, Y. 2014.
\newblock Generative Adversarial Nets.
\newblock In Ghahramani, Z.; Welling, M.; Cortes, C.; Lawrence, N.; and
  Weinberger, K.~Q., eds., \emph{Advances in Neural Information Processing
  Systems}, volume~27. Curran Associates, Inc.

\bibitem[{Harms et~al.(2020)Harms, Lei, Wang, Mcdonald, Ghavidel, and
  Stokes}]{Harms2020}
Harms, J.; Lei, Y.; Wang, T.; Mcdonald, M.; Ghavidel, B.; and Stokes, W. 2020.
\newblock {Cone-beam CT-derived relative stopping power map generation via deep
  learning for proton radiotherapy}.
\newblock 47(9): 4416--4427.

\bibitem[{Hassani et~al.(2021)Hassani, Walton, Shah, Abuduweili, Li, and
  Shi}]{Hassani2021}
Hassani, A.; Walton, S.; Shah, N.; Abuduweili, A.; Li, J.; and Shi, H. 2021.
\newblock {Escaping the Big Data Paradigm with Compact Transformers}.

\bibitem[{Hendrycks and Gimpel(2016)}]{Hendrycks2016}
Hendrycks, D.; and Gimpel, K. 2016.
\newblock {Gaussian Error Linear Units (GELUs)}.
\newblock 1--9.

\bibitem[{Hussein et~al.(2018)Hussein, Heijmen, Verellen, and
  Nisbet}]{Hussein2018}
Hussein, M.; Heijmen, B. J.~M.; Verellen, D.; and Nisbet, A. 2018.
\newblock {Automation in intensity modulated radiotherapy treatment
  planning—a review of recent innovations}.
\newblock \emph{The British Journal of Radiology}, 91(1092): 20180270.

\bibitem[{Javaid et~al.(2021)Javaid, Souris, Huang, and Lee}]{Javaid2021}
Javaid, U.; Souris, K.; Huang, S.; and Lee, J.~A. 2021.
\newblock {Denoising proton therapy Monte Carlo dose distributions in multiple
  tumor sites: A comparative neural networks architecture study}.
\newblock \emph{Physica Medica}, 89(August): 93--103.

\bibitem[{Kajikawa et~al.(2019)Kajikawa, Kadoya, Ito, Takayama, Chiba, Tomori,
  Nemoto, Dobashi, Takeda, and Jingu}]{Kajikawa2019}
Kajikawa, T.; Kadoya, N.; Ito, K.; Takayama, Y.; Chiba, T.; Tomori, S.; Nemoto,
  H.; Dobashi, S.; Takeda, K.; and Jingu, K. 2019.
\newblock {A convolutional neural network approach for IMRT dose distribution
  prediction in prostate cancer patients}.
\newblock \emph{Journal of Radiation Research}, 60(5): 685--693.

\bibitem[{Kearney et~al.(2018)Kearney, Chan, Haaf, Descovich, and
  Solberg}]{Kearney2018}
Kearney, V.; Chan, J.~W.; Haaf, S.; Descovich, M.; and Solberg, T.~D. 2018.
\newblock {DoseNet: a volumetric dose prediction algorithm using 3D
  fully-convolutional neural networks}.
\newblock \emph{Physics in Medicine \& Biology}, 63(23): 235022.

\bibitem[{Kontaxis et~al.(2020)Kontaxis, Bol, Lagendijk, and
  Raaymakers}]{Kontaxis2020}
Kontaxis, C.; Bol, G.~H.; Lagendijk, J. J.~W.; and Raaymakers, B.~W. 2020.
\newblock {DeepDose: Towards a fast dose calculation engine for radiation
  therapy using deep learning}.
\newblock \emph{Physics in Medicine \& Biology}, 65(7): 075013.

\bibitem[{Lalonde et~al.(2020)Lalonde, Winey, Verburg, Paganetti, and
  Sharp}]{Lalonde2020}
Lalonde, A.; Winey, B.; Verburg, J.; Paganetti, H.; and Sharp, G.~C. 2020.
\newblock {Evaluation of CBCT scatter correction using deep convolutional
  neural networks for head and neck adaptive proton therapy}.
\newblock \emph{Physics in Medicine \& Biology Biology}, 65(24): 245022.

\bibitem[{Lee et~al.(2019)Lee, Kim, Kwak, Kim, Lee, Cho, and Cho}]{Lee2019}
Lee, H.; Kim, H.; Kwak, J.; Kim, Y.~S.; Lee, S.~W.; Cho, S.; and Cho, B. 2019.
\newblock {Fluence-map generation for prostate intensity-modulated radiotherapy
  planning using a deep-neural-network}.
\newblock \emph{Scientific Reports}, 9(1): 15671.

\bibitem[{Low et~al.(1998)Low, Harms, Mutic, and Purdy}]{Low1998}
Low, D.~A.; Harms, W.~B.; Mutic, S.; and Purdy, J.~A. 1998.
\newblock {A technique for the quantitative evaluation of dose distributions}.
\newblock \emph{Medical Physics}, 25(5): 656--661.

\bibitem[{Lundkvist et~al.(2005)Lundkvist, Ekman, Ericsson, Jönsson, and
  Glimelius}]{Lundkvist2005}
Lundkvist, J.; Ekman, M.; Ericsson, S.~R.; Jönsson, B.; and Glimelius, B.
  2005.
\newblock Proton therapy of cancer: Potential clinical advantages and
  cost-effectiveness.
\newblock \emph{Acta Oncologica}, 44(8): 850--861.

\bibitem[{Meyer et~al.(2018)Meyer, Noblet, Mazzara, and Lallement}]{Meyer2018}
Meyer, P.; Noblet, V.; Mazzara, C.; and Lallement, A. 2018.
\newblock {Survey on deep learning for radiotherapy}.
\newblock \emph{Computers in Biology and Medicine}, 98(May): 126--146.

\bibitem[{Neishabouri et~al.(2021)Neishabouri, Wahl, Mairani, K{\"{o}}the, and
  Bangert}]{Neishabouri2021}
Neishabouri, A.; Wahl, N.; Mairani, A.; K{\"{o}}the, U.; and Bangert, M. 2021.
\newblock {Long short-term memory networks for proton dose calculation in
  highly heterogeneous tissues}.
\newblock \emph{Medical Physics}, 48(4): 1893--1908.

\bibitem[{Neph et~al.(2021)Neph, Lyu, Huang, Yang, and Sheng}]{Neph2021}
Neph, R.; Lyu, Q.; Huang, Y.; Yang, Y.~M.; and Sheng, K. 2021.
\newblock {DeepMC: a deep learning method for efficient Monte Carlo beamlet
  dose calculation by predictive denoising in magnetic resonance-guided
  radiotherapy}.
\newblock \emph{Physics in Medicine \& Biology}, 66(3): 035022.

\bibitem[{Nguyen et~al.(2019{\natexlab{a}})Nguyen, Jia, Sher, Lin, Iqbal, Liu,
  and Jiang}]{Nguyen2019a}
Nguyen, D.; Jia, X.; Sher, D.; Lin, M.-H.; Iqbal, Z.; Liu, H.; and Jiang, S.
  2019{\natexlab{a}}.
\newblock {3D radiotherapy dose prediction on head and neck cancer patients
  with a hierarchically densely connected U-net deep learning architecture}.
\newblock \emph{Physics in Medicine \& Biology}, 64(6): 065020.

\bibitem[{Nguyen et~al.(2019{\natexlab{b}})Nguyen, Long, Jia, Lu, Gu, Iqbal,
  and Jiang}]{Nguyen2019}
Nguyen, D.; Long, T.; Jia, X.; Lu, W.; Gu, X.; Iqbal, Z.; and Jiang, S.
  2019{\natexlab{b}}.
\newblock {A feasibility study for predicting optimal radiation therapy dose
  distributions of prostate cancer patients from patient anatomy using deep
  learning}.
\newblock \emph{Scientific Reports}, 9(1): 1076.

\bibitem[{Peng et~al.(2019{\natexlab{a}})Peng, Shan, Liu, Pei, Wang, and
  Xu}]{Peng2019}
Peng, Z.; Shan, H.; Liu, T.; Pei, X.; Wang, G.; and Xu, X.~G.
  2019{\natexlab{a}}.
\newblock {MCDNet – A Denoising Convolutional Neural Network to Accelerate
  Monte Carlo Radiation Transport Simulations: A Proof of Principle With
  Patient Dose From X-Ray CT Imaging}.
\newblock \emph{IEEE Access}, 7: 76680--76689.

\bibitem[{Peng et~al.(2019{\natexlab{b}})Peng, Shan, Liu, Pei, Zhou, Wang, and
  Xu}]{Peng2019a}
Peng, Z.; Shan, H.; Liu, T.; Pei, X.; Zhou, J.; Wang, G.; and Xu, X.~G.
  2019{\natexlab{b}}.
\newblock {Deep learning for accelerating Monte Carlo radiation transport
  simulation in intensity-modulated radiation therapy}.
\newblock 1--8.

\bibitem[{Pereira, Traughber, and Muzic(2014)}]{Pereira2014}
Pereira, G.~C.; Traughber, M.; and Muzic, R.~F. 2014.
\newblock {The Role of Imaging in Radiation Therapy Planning: Past, Present,
  and Future}.
\newblock \emph{BioMed Research International}, 2014(2): 1--9.

\bibitem[{Ramachandran et~al.(2019)Ramachandran, Bello, Parmar, Levskaya,
  Vaswani, and Shlens}]{Ramachandran2019}
Ramachandran, P.; Bello, I.; Parmar, N.; Levskaya, A.; Vaswani, A.; and Shlens,
  J. 2019.
\newblock {Stand-alone self-attention in vision models}.
\newblock \emph{Advances in Neural Information Processing Systems}, 32.

\bibitem[{Ronneberger, Fischer, and Brox(2015)}]{Ronneberger2015}
Ronneberger, O.; Fischer, P.; and Brox, T. 2015.
\newblock U-Net: Convolutional Networks for Biomedical Image Segmentation.
\newblock In Navab, N.; Hornegger, J.; Wells, W.~M.; and Frangi, A.~F., eds.,
  \emph{Medical Image Computing and Computer-Assisted Intervention -- MICCAI
  2015}, 234--241. Cham: Springer International Publishing.
\newblock ISBN 978-3-319-24574-4.

\bibitem[{Souris, Lee, and Sterpin(2016)}]{Souris2016}
Souris, K.; Lee, J.~A.; and Sterpin, E. 2016.
\newblock {Fast multipurpose Monte Carlo simulation for proton therapy using
  multi- and many-core CPU architectures}.
\newblock \emph{Medical Physics}, 43(4): 1700--1712.

\bibitem[{Srivastava et~al.(2014)Srivastava, Hinton, Krizhevsky, and
  Salakhutdinov}]{Srivastava2014}
Srivastava, N.; Hinton, G.; Krizhevsky, A.; and Salakhutdinov, R. 2014.
\newblock {Dropout: A Simple Way to Prevent Neural Networks from Overfitting}.
\newblock \emph{Journal of Machine Learning Research}, 15: 1929--1958.

\bibitem[{Sung et~al.(2021)Sung, Ferlay, Siegel, Laversanne, Soerjomataram,
  Jemal, and Bray}]{Sung2021}
Sung, H.; Ferlay, J.; Siegel, R.~L.; Laversanne, M.; Soerjomataram, I.; Jemal,
  A.; and Bray, F. 2021.
\newblock Global Cancer Statistics 2020: GLOBOCAN Estimates of Incidence and
  Mortality Worldwide for 36 Cancers in 185 Countries.
\newblock \emph{CA: A Cancer Journal for Clinicians}, 71(3): 209--249.

\bibitem[{Touvron et~al.(2020)Touvron, Cord, Douze, Massa, Sablayrolles, and
  J{\'{e}}gou}]{Touvron2020}
Touvron, H.; Cord, M.; Douze, M.; Massa, F.; Sablayrolles, A.; and J{\'{e}}gou,
  H. 2020.
\newblock {Training data-efficient image transformers \& distillation through
  attention}.
\newblock 1--22.

\bibitem[{Tsekas et~al.(2021)Tsekas, Bol, Raaymakers, and
  Kontaxis}]{Tsekas2021}
Tsekas, G.; Bol, G.~H.; Raaymakers, B.~W.; and Kontaxis, C. 2021.
\newblock {DeepDose: a robust deep learning-based dose engine for abdominal
  tumours in a 1.5 T MRI radiotherapy system}.
\newblock \emph{Physics in Medicine \& Biology}, 66(6): 065017.

\bibitem[{Vaswani et~al.(2017)Vaswani, Shazeer, Parmar, Uszkoreit, Jones,
  Gomez, Kaiser, and Polosukhin}]{Vaswani2017}
Vaswani, A.; Shazeer, N.; Parmar, N.; Uszkoreit, J.; Jones, L.; Gomez, A.~N.;
  Kaiser, {\L}.; and Polosukhin, I. 2017.
\newblock {Attention is all you need}.
\newblock In \emph{Advances in Neural Information Processing Systems}, volume
  2017-Decem, 5999--6009.

\bibitem[{Wang et~al.(2020)Wang, Sheng, Wang, Zhang, Li, Palta, Czito, Willett,
  Wu, Ge, Yin, and Wu}]{Wang2020}
Wang, W.; Sheng, Y.; Wang, C.; Zhang, J.; Li, X.; Palta, M.; Czito, B.;
  Willett, C.~G.; Wu, Q.; Ge, Y.; Yin, F.-F.; and Wu, Q.~J. 2020.
\newblock {Fluence Map Prediction Using Deep Learning Models – Direct Plan
  Generation for Pancreas Stereotactic Body Radiation Therapy}.
\newblock \emph{Frontiers in Artificial Intelligence}, 3(September): 1--10.

\bibitem[{Wieser et~al.(2017)Wieser, Cisternas, Wahl, Ulrich, Stadler, Mescher,
  Muller, Klinge, Gabrys, Burigo, Mairani, Ecker, Ackermann, Ellerbrock,
  Parodi, Jakel, and Bangert}]{Wieser2017}
Wieser, H.~P.; Cisternas, E.; Wahl, N.; Ulrich, S.; Stadler, A.; Mescher, H.;
  Muller, L.~R.; Klinge, T.; Gabrys, H.; Burigo, L.; Mairani, A.; Ecker, S.;
  Ackermann, B.; Ellerbrock, M.; Parodi, K.; Jakel, O.; and Bangert, M. 2017.
\newblock {Development of the open-source dose calculation and optimization
  toolkit matRad}.
\newblock \emph{Medical Physics}, 44(6): 2556--2568.

\bibitem[{Wu et~al.(2021)Wu, Nguyen, Xing, Montero, Schuemann, Shang, Pu, and
  Jiang}]{Wu2021}
Wu, C.; Nguyen, D.; Xing, Y.; Montero, A.~B.; Schuemann, J.; Shang, H.; Pu, Y.;
  and Jiang, S. 2021.
\newblock {Improving proton dose calculation accuracy by using deep learning}.
\newblock \emph{Machine Learning: Science and Technology}, 2(1): 015017.

\bibitem[{Wu and He(2020)}]{Wu2020}
Wu, Y.; and He, K. 2020.
\newblock {Group Normalization}.
\newblock \emph{International Journal of Computer Vision}, 128(3): 742--755.

\bibitem[{Xing et~al.(2020{\natexlab{a}})Xing, Nguyen, Lu, Yang, and
  Jiang}]{Xing2020}
Xing, Y.; Nguyen, D.; Lu, W.; Yang, M.; and Jiang, S. 2020{\natexlab{a}}.
\newblock {Technical Note: A feasibility study on deep learning‐based
  radiotherapy dose calculation}.
\newblock \emph{Medical Physics}, 47(2): 753--758.

\bibitem[{Xing et~al.(2020{\natexlab{b}})Xing, Zhang, Nguyen, Lin, Lu, and
  Jiang}]{Xing2020a}
Xing, Y.; Zhang, Y.; Nguyen, D.; Lin, M.; Lu, W.; and Jiang, S.
  2020{\natexlab{b}}.
\newblock {Boosting radiotherapy dose calculation accuracy with deep learning}.
\newblock \emph{Journal of Applied Clinical Medical Physics}, 21(8): 149--159.

\bibitem[{You et~al.(2019)You, Li, Reddi, Hseu, Kumar, Bhojanapalli, Song,
  Demmel, Keutzer, and Hsieh}]{You2019}
You, Y.; Li, J.; Reddi, S.; Hseu, J.; Kumar, S.; Bhojanapalli, S.; Song, X.;
  Demmel, J.; Keutzer, K.; and Hsieh, C.-J. 2019.
\newblock {Large Batch Optimization for Deep Learning: Training BERT in 76
  minutes}.

\bibitem[{Zhang et~al.(2021)Zhang, Yue, Su, Liu, Ding, Zhou, Wang, Kuang, and
  Nie}]{Zhang2021}
Zhang, Y.; Yue, N.; Su, M.-y.; Liu, B.; Ding, Y.; Zhou, Y.; Wang, H.; Kuang,
  Y.; and Nie, K. 2021.
\newblock {Improving CBCT quality to CT level using deep learning with
  generative adversarial network}.
\newblock \emph{Medical Physics}, 48(June): 2816--2826.

\bibitem[{Zhu, Liu, and Chen(2020)}]{Zhu2020}
Zhu, J.; Liu, X.; and Chen, L. 2020.
\newblock {A preliminary study of a photon dose calculation algorithm using a
  convolutional neural network}.
\newblock \emph{Physics in Medicine \& Biology}, 65(20): 20NT02.

\end{thebibliography}
\normalsize

\newpage
\appendix
\section{Model hyperparameters}
\label{app:hyperparameter}
While the dimension of the inputs and outputs is fixed, the different choices of model hyperparameters offer great flexibility in the design DoTA's architecture and have an effect on the final accuracy. To build the best performing algorithm, we experiment with varying the number of transformer blocks $N$, the number of filters $K$ after the last layer in the convolutional encoder, and the number of attention heads $N_h$. Given the internal memory limitations of the GPU used in our experiments, we only report combinations that are compatible with a mini-batch size of 8 samples. \tabref{hs} shows the results of the hyperparameter search, showing the performance of all models on the test set, highlighting that the final combination of hyperparameters chosen based on the lowest validation set error indeed yields the lowest test set error, showcasing good generalization of the model.

\begin{table}[h!]
	\centering
	\caption{Model hyperparameter tuning experiment. A separate model is trained for each combination of transformer blocks $N$, convolutional filters $K$ and attention heads $N_h$, using the same learning rate and batch size.}
	\begin{tabular}{@{}cccc@{}}
		\toprule
		\textbf{\begin{tabular}[c]{@{}c@{}}Transformer\\  blocks $N$\end{tabular}} & \textbf{\begin{tabular}[c]{@{}c@{}}Convolution\\  filters $K$\end{tabular}} & \textbf{\begin{tabular}[c]{@{}c@{}}Attention \\ heads $N_h$\end{tabular}} & \begin{tabular}[c]{@{}c@{}}\textbf{Test MSE} \\ {[}Gy/$10^9$ part.$]^2$\end{tabular} \\ \midrule
		\textbf{1} & \textbf{10} & \textbf{16} & \textbf{0.0192} \\
		1 & 16 & 16 & 0.0263 \\
		2 & 8 & 8 & 0.0560 \\
		2 & 8 & 16 & 0.0451 \\
		2 & 10 & 8 & 0.0273 \\
		2 & 16 & 8 & 0.0466 \\
		2 & 16 & 16 & 0.0396 \\
		4 & 8 & 8 & 0.0843 \\ \bottomrule
	\end{tabular}
	\label{tab:hs}
\end{table}

\section{Additional test geometries}
\label{app:geometries}
\begin{figure}[]
	\centering
	\includegraphics[width=0.49\textwidth]{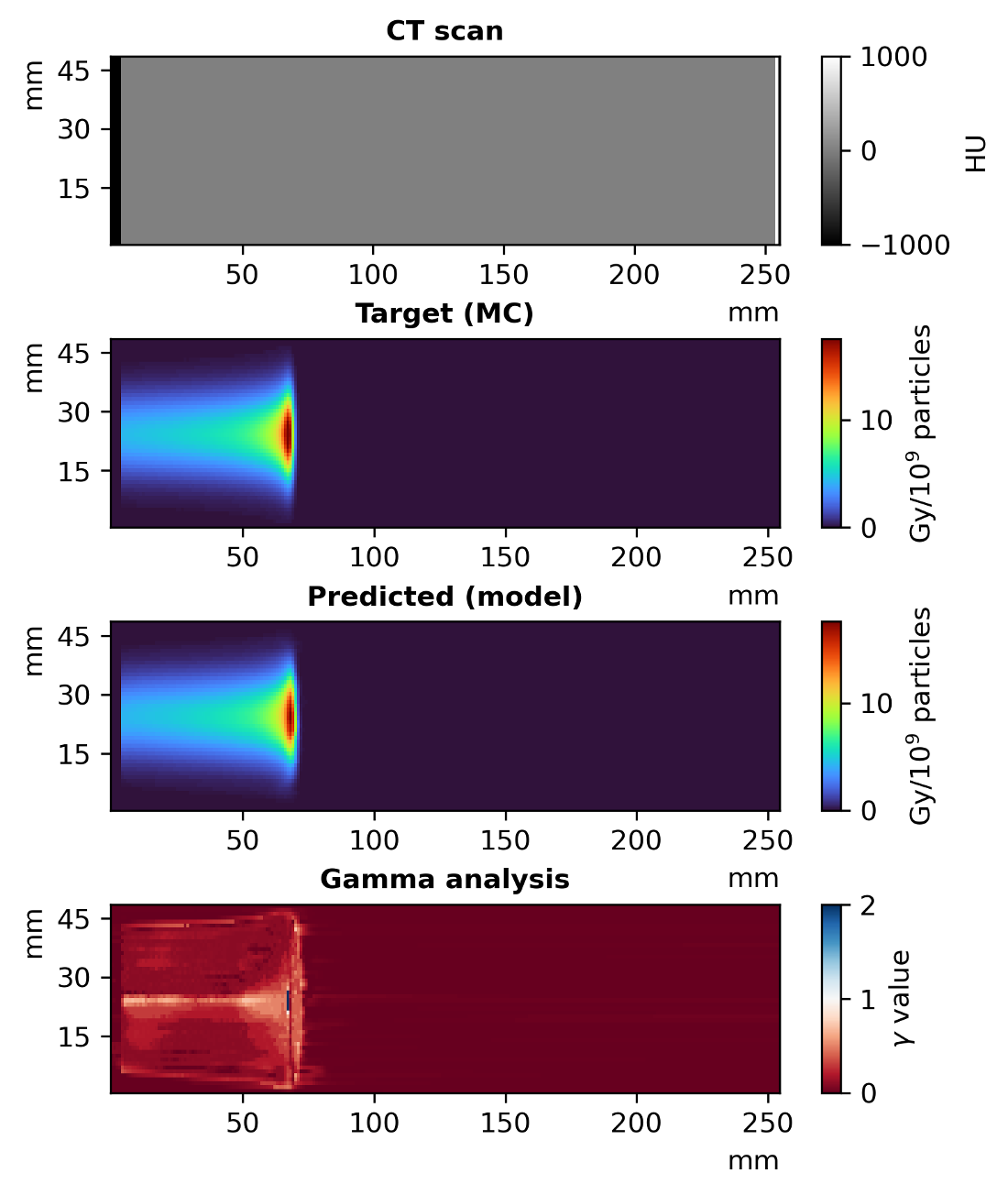} 
	\caption{DoTA's predicted dose distribution in a uniform volume of water for a 90 MeV mono-energetic proton beam.}
	\label{fig:water}
\end{figure}

\begin{figure*}[]
	\centering
	\begin{subfigure}[]{0.49\textwidth}
		\centering
		\includegraphics[width=\textwidth]{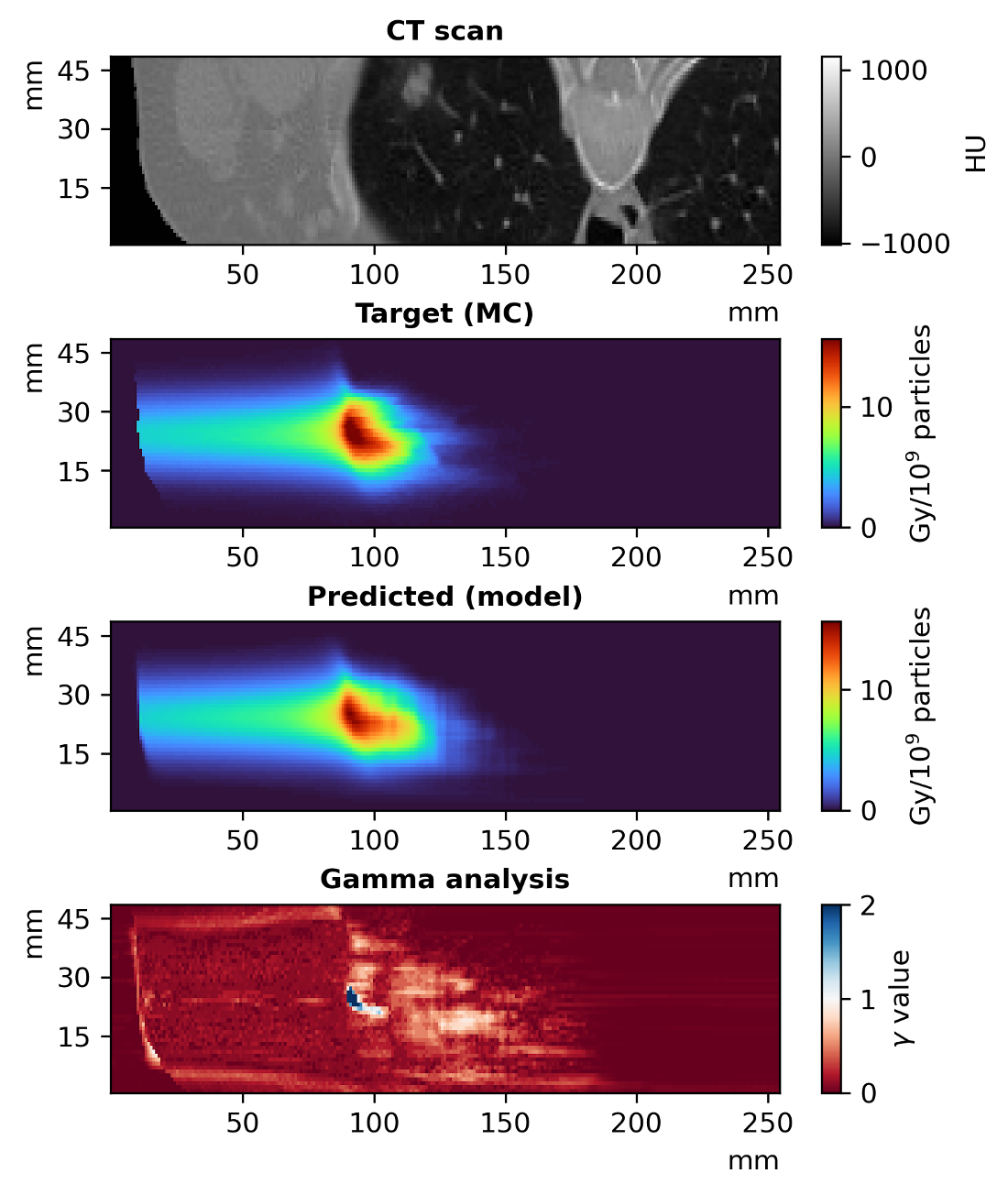}
	\end{subfigure}
	\begin{subfigure}[]{0.49\textwidth}
		\centering
		\includegraphics[width=\textwidth]{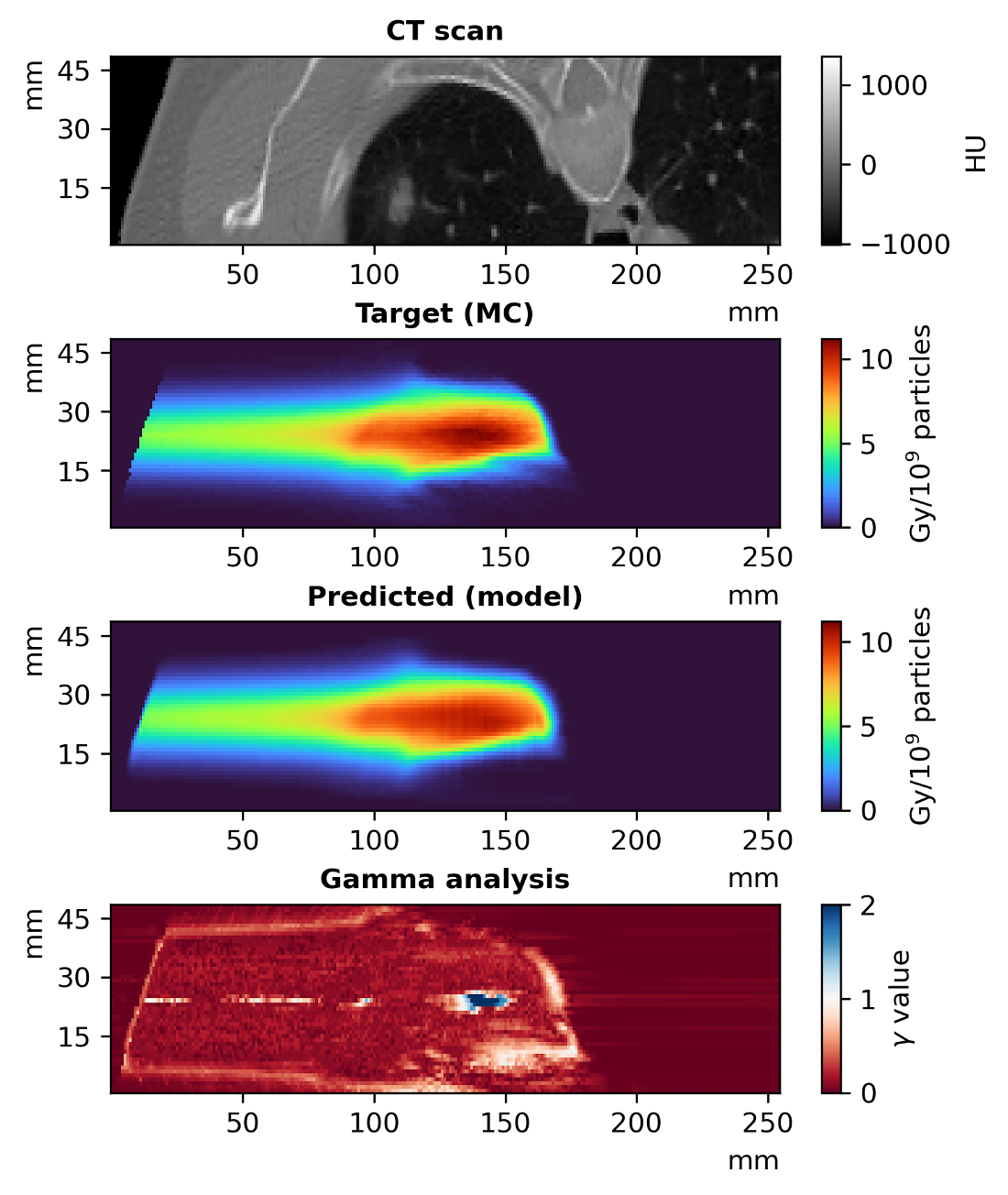}
	\end{subfigure}
	\caption{Predicted dose distribution and gamma values for the test samples corresponding to the 2 gamma pass rates closest to the median gamma pass rate.}
	\label{fig:median}
\end{figure*}

\figref{median} shows the accuracy that DoTA achieves in the two samples corresponding to the 2 gamma pass rates closest to the median gamma pass rate among the test set samples (99.5915\% and 99.592\% for the left and right of \figref{fig:median}, respectively). The predicted dose distributions closely follow the Monte Carlo ground truth, with only a handful of voxels being inaccurate. Additionally, to verify that the model learns the basic proton physics beyond lung patient geometries, we evaluate DoTA in a series of homogeneous volumes that are unseen during training. \figref{water} displays one such prediction for a 90 MeV mono-energetic pencil beam travelling through a water phantom, clearly demonstrating that the model implicitly captures physics and generalizes very well to simple, but unseen geometries entirely different from the training samples.

\end{document}